\documentclass[conf]{new-aiaa}
\usepackage[utf8]{inputenc}

\usepackage{graphicx}
\usepackage{amsmath}
\usepackage{amssymb}
\usepackage{algorithm}
\usepackage{algorithmic}
\usepackage{booktabs}
\usepackage{siunitx}
\usepackage{longtable,tabularx}
\usepackage{url}
\usepackage{hyperref}
\usepackage{tikz}
\usetikzlibrary{shapes.geometric, arrows.meta, positioning, calc, fit, backgrounds, decorations.pathreplacing, shadows, patterns}

\definecolor{inputcolor}{RGB}{66, 133, 244}
\definecolor{encodercolor}{RGB}{52, 168, 83}
\definecolor{reasoningcolor}{RGB}{251, 188, 4}
\definecolor{decodercolor}{RGB}{234, 67, 53}
\definecolor{outputcolor}{RGB}{156, 39, 176}
\definecolor{feedbackcolor}{RGB}{255, 87, 34}
\definecolor{latentcolor}{RGB}{0, 150, 136}
\definecolor{lightgray}{RGB}{245, 245, 245}
\definecolor{darkgray}{RGB}{97, 97, 97}

\usepackage{multirow} 
\title{Tiny Recursive Control: Iterative Reasoning for Efficient Optimal Control}

\author{Amit Jain\footnote{Postdoctoral Associate, Department of Aeronautics and Astronautics,  Email: amitjain@mit.edu.}}
\affil{Massachusetts Institute of Technology, Cambridge, MA 02139}
\author{Richard Linares\footnote{Associate Professor, Department of Aeronautics and Astronautics,  Email: linaresr@mit.edu.}}
\affil{Massachusetts Institute of Technology, Cambridge, MA 02139}

\begin{document}

\maketitle

% ==============================================================================
% ABSTRACT
% ==============================================================================

\begin{abstract}
Neural network controllers increasingly demand millions of parameters, and language model approaches push into the billions. For embedded aerospace systems with strict power and latency constraints, this scaling is prohibitive. We present Tiny Recursive Control (TRC), a neural architecture based on a counterintuitive principle: capacity can emerge from iteration depth rather than parameter count. TRC applies compact networks (approximately 1.5M parameters) repeatedly through a two-level hierarchical latent structure, refining control sequences by simulating trajectories and correcting based on tracking error. Because the same weights process every refinement step, adding iterations increases computation without increasing memory. We evaluate TRC on nonlinear control problems including oscillator stabilization and powered descent with fuel constraints. Across these domains, TRC achieves near-optimal control costs while requiring only millisecond-scale inference on GPU and under 10~MB memory, two orders of magnitude smaller than language model baselines. These results demonstrate that recursive reasoning, previously confined to discrete tasks, transfers effectively to continuous control synthesis.
\end{abstract}

% ==============================================================================
% NOMENCLATURE
% ==============================================================================

\section*{Nomenclature}

{\renewcommand\arraystretch{1.0}
\noindent\begin{longtable*}{@{}l @{\quad=\quad} l@{}}
\multicolumn{2}{@{}l}{\textit{State and Control Variables}} \\
$\mathbf{x}$    & state vector \\
$\mathbf{u}$    & control input vector \\
$\mathbf{e}$    & trajectory error vector \\
$\Delta\mathbf{u}$ & control residual update \\
$\mathbf{x}_{\text{target}}$ & target state \\
$\mathbf{u}^*$  & optimal control sequence (ground truth) \\[0.5em]
\multicolumn{2}{@{}l}{\textit{Architecture Components}} \\
$\mathbf{z}_H$  & high-level (strategic) latent state \\
$\mathbf{z}_L$  & low-level (tactical) latent state \\
$\mathcal{L}_\theta$ & shared reasoning module \\
$\mathbf{H}_{\text{init}}$ & learnable high-level initialization \\
$\mathbf{L}_{\text{init}}$ & learnable low-level initialization \\[0.5em]
\multicolumn{2}{@{}l}{\textit{Dimensions}} \\
$d_x$           & state dimension \\
$d_u$           & control dimension \\
$d_z$           & latent dimension \\
$d_h$           & hidden dimension \\[0.5em]
\multicolumn{2}{@{}l}{\textit{Iterations and Horizon}} \\
$K$             & number of high-level refinement iterations \\
$n$             & number of low-level reasoning cycles per iteration \\
$L$             & number of reasoning blocks per cycle \\
$T$             & control horizon (number of time steps) \\[0.5em]
\multicolumn{2}{@{}l}{\textit{System Matrices}} \\
$\mathbf{A}$    & state transition matrix \\
$\mathbf{B}$    & control input matrix \\
$\mathbf{Q}$    & state cost matrix \\
$\mathbf{R}$    & control cost matrix \\
$\mathbf{Q}_f$  & terminal state cost matrix \\[0.5em]
\multicolumn{2}{@{}l}{\textit{Cost and Training}} \\
$J$             & cost function \\
$\ell$, $\ell_f$ & running and terminal cost functions \\
$f$             & dynamics function (state transition map) \\
$\lambda$       & process supervision weight \\
$\eta$          & learning rate / step size \\
$B$             & training batch size \\[0.5em]
\multicolumn{2}{@{}l}{\textit{Problem-Specific Parameters}} \\
$\mu$           & Van der Pol damping parameter \\
$I_{\text{sp}}$ & specific impulse \\
$g_0$           & Earth's gravitational acceleration (9.81 m/s$^2$)
\end{longtable*}}

% ==============================================================================
% INTRODUCTION
% ==============================================================================

\section{Introduction}

\lettrine{N}{eural} networks have become standard tools for control synthesis, yet their computational requirements continue to grow. Transformer-based controllers now routinely exceed tens of millions of parameters; language model approaches push into the billions. For problems where classical methods like LQR solve optimally in microseconds, this scaling raises a natural question: how much of this capacity is necessary, and how much is architectural overhead? Embedded systems cannot absorb the overhead. Satellite autonomy, UAV swarms, and reusable launch vehicles all demand controllers that fit within power budgets measured in watts and latencies measured in milliseconds. A rocket guidance computer cannot host a 7-billion-parameter model. An attitude control loop running at 100 Hz cannot wait 500 ms for inference. The gap between research demonstrations and deployable controllers remains wide precisely because model efficiency has received less attention than model capability.

This computational mismatch has deeper roots than parameter count alone. Standard feedforward networks lack the iterative structure that optimal control inherently requires. They must ``memorize'' solutions rather than ``compute'' them. A simple oscillator and a powered descent vehicle receive the same computational budget, regardless of problem difficulty. The question is not ``how small can we make the network?'' but rather ``can we decouple capacity from parameter count?''

Control synthesis spans a computational spectrum. Analytical methods like LQR compute optimal feedback gains directly from system matrices~\cite{anderson1971linear,lewis2012optimal}. For linear dynamics with quadratic costs, the Riccati equation yields the solution in closed form, completing in microseconds. Nonlinear and constrained problems require numerical optimization: Model Predictive Control solves finite-horizon problems at each timestep~\cite{camacho2013model,rawlings2017model}, with solution times ranging from milliseconds to seconds. Prior work has explored various formulations for optimal feedback synthesis, including dynamic programming~\cite{bertsekas2012dynamic}, direct collocation~\cite{betts1998survey}, and Hamilton-Jacobi approaches~\cite{jain2025hamilton,jain2023sparse}, yet the fundamental tradeoff between solution quality and computational cost persists. Between these extremes, neural approximations trade optimality guarantees for speed.

Neural network approximations shift computation from online optimization to offline training. Chen et al.~\cite{chen2018approximating} and Hertneck et al.~\cite{hertneck2018learning} trained feedforward networks to approximate MPC policies. Wu et al.~\cite{wu2024composing} combined neural predictions with online refinement. Celestini et al.~\cite{celestini2024transformer} used transformers for warm-start solutions. Jain et al.~\cite{jain2025multi} applied transformer-based reinforcement learning to multi-phase spacecraft trajectories, demonstrating that attention mechanisms can capture long-horizon dependencies in sequential decision problems. More recently, language models have generated control sequences for spacecraft systems~\cite{brohan2023rt2,zucchelli2025finetuned,carrasco2024finetuning}. Imitation learning offers another path: behavior cloning learns policies directly from demonstrations~\cite{pomerleau1991efficient,posadas2025action}, while reinforcement learning approaches like PPO~\cite{schulman2017proximal} and SAC~\cite{haarnoja2018soft} learn through environment interaction. These methods achieve 10-100$\times$ speedups over MPC, but share three fundamental limitations: (1) model size remains fixed regardless of problem difficulty, so a low-dimensional oscillator and a 6-DOF rocket use the same multi-million parameter network; (2) errors compound without correction, analogous to open-loop control where the policy has no opportunity to observe and correct its mistakes; and (3) the computational cost is set at design time, not adapted to the task at hand.

A different philosophy asks: what if computation could scale with problem difficulty? Iterative refinement architectures offer one answer. In computer vision, detection models like DETR refine predictions through repeated self-attention~\cite{carion2020end}; in 3D reconstruction, R2N2 produces updates resembling numerical solvers~\cite{seeliger2022r2n2}. The common thread is that output quality improves with iteration count, not parameter count. The clearest demonstration comes from Tiny Recursive Models (TRM) for discrete reasoning~\cite{jolicoeurmartineau2024lessmore}. The result is counterintuitive: TRM's 7-million-parameter network (0.01\% the size of competing 70-billion-parameter models) achieves comparable reasoning accuracy. The key is weight sharing: the same network weights apply across $K$ refinement steps, each containing $n$ internal reasoning cycles. The network learns a refinement \textit{operator}, not a direct input-output mapping. For iterative problems, capacity emerges from computation depth, not parameter count. This principle has not been applied to continuous optimal control.

This paper develops \textit{Tiny Recursive Control} (TRC), importing the ``less is more'' principle to continuous control. The core insight: a small network applied repeatedly can match a large network applied once, while using constant memory. TRC generates an initial control estimate, simulates the resulting trajectory through known dynamics, and refines based on tracking error, with the same compact network (approximately 1.5M parameters) processing every iteration. Trained on optimal trajectories, TRC achieves near-optimal control costs on Van der Pol oscillator and powered descent problems while requiring millisecond-scale inference on GPU and under 10 MB memory, two orders of magnitude smaller than language model baselines. To our knowledge, this is the first application of recursive weight-shared architectures to continuous-time optimal control.

Section~\ref{sec:problem} formulates the optimal control problem. Section~\ref{sec:method} presents the TRC architecture and its theoretical foundation. Section~\ref{sec:training} describes the training methodology. Section~\ref{sec:experiments} reports experimental results. Section~\ref{sec:conclusion} discusses implications, limitations, and conclusions.

% ==============================================================================
% PROBLEM FORMULATION
% ==============================================================================

\section{Problem Formulation}
\label{sec:problem}

This work addresses finite-horizon optimal control problems with terminal cost penalties, a formulation that captures common aerospace tasks such as rendezvous maneuvers, landing trajectories, and spacecraft detumbling. In each case, the objective is to reach a specified target state while minimizing control effort over a fixed time horizon. The finite-horizon structure enables supervised learning from optimal demonstrations, and the problem admits a natural iterative refinement interpretation that TRC exploits.

The system under control evolves in discrete time according to state $\mathbf{x}_t \in \mathcal{X} \subseteq \mathbb{R}^{d_x}$ and control input $\mathbf{u}_t \in \mathcal{U} \subseteq \mathbb{R}^{d_u}$, with dynamics governed by the transition map $f: \mathcal{X} \times \mathcal{U} \to \mathcal{X}$:
\begin{equation}
\label{eq:dynamics}
\mathbf{x}_{t+1} = f(\mathbf{x}_t, \mathbf{u}_t), \quad t = 0, 1, \ldots, T-1
\end{equation}
The transition map $f$ is assumed continuously differentiable, enabling gradient-based optimization during training. Control inputs must lie within an admissible set defined by actuator limits, expressed as the compact box $\mathcal{U} = \{\mathbf{u} \in \mathbb{R}^{d_u} : u_{\min} \leq \mathbf{u} \leq u_{\max}\}$ where inequalities hold element-wise. This constraint captures practical limitations common across aerospace systems, including thrust bounds and control surface deflection limits.

Given an initial state $\mathbf{x}_0$ and target state $\mathbf{x}_{\text{target}}$, the control synthesis task seeks a sequence $\mathbf{u}_{0:T-1} = (\mathbf{u}_0, \ldots, \mathbf{u}_{T-1})$ that solves the following optimization problem:
\begin{equation}
\label{eq:ocp}
\begin{aligned}
\min_{\mathbf{u}_{0:T-1}} \quad & J(\mathbf{u}_{0:T-1}) = \sum_{t=0}^{T-1} \ell(\mathbf{x}_t, \mathbf{u}_t) + \ell_f(\mathbf{x}_T, \mathbf{x}_{\text{target}}) \\
\text{subject to} \quad & \mathbf{x}_{t+1} = f(\mathbf{x}_t, \mathbf{u}_t), \quad t = 0, \ldots, T-1 \\
& \mathbf{u}_t \in \mathcal{U}, \quad t = 0, \ldots, T-1 \\
& \mathbf{x}_0 \text{ given}
\end{aligned}
\end{equation}
The cost function balances two objectives through the running cost $\ell: \mathcal{X} \times \mathcal{U} \to \mathbb{R}_{\geq 0}$, which penalizes state deviation and control effort along the trajectory, and the terminal cost $\ell_f: \mathcal{X} \times \mathcal{X} \to \mathbb{R}_{\geq 0}$, which penalizes deviation from the target at the final time. For the quadratic formulation considered in this work, these take the form:
\begin{equation}
\label{eq:cost}
\ell(\mathbf{x}, \mathbf{u}) = \mathbf{x}^\top \mathbf{Q} \mathbf{x} + \mathbf{u}^\top \mathbf{R} \mathbf{u}, \quad \ell_f(\mathbf{x}_T, \mathbf{x}_{\text{target}}) = (\mathbf{x}_T - \mathbf{x}_{\text{target}})^\top \mathbf{Q}_f (\mathbf{x}_T - \mathbf{x}_{\text{target}})
\end{equation}
where $\mathbf{Q} \succeq 0$, $\mathbf{R} \succ 0$, and $\mathbf{Q}_f \succeq 0$ are symmetric weight matrices. Setting $\mathbf{Q} = 0$ recovers the fuel-optimal formulation used in powered descent, while nonzero $\mathbf{Q}$ enables state regulation as in the oscillator problem. TRC learns to approximate optimal control sequences for this class of problems, handling both linear and nonlinear dynamics through supervised learning on optimal demonstrations.

Classical algorithms for solving~\eqref{eq:ocp} share a common iterative pattern that motivates the TRC architecture: initialize a control sequence, simulate the resulting trajectory through the dynamics, observe the deviation from the target, and compute a correction. This cycle repeats until convergence, with each refinement step performing the same fundamental operation (mapping trajectory error to control adjustment) regardless of iteration count. TRC exploits this structure by learning a refinement operator rather than a direct mapping from states to controls. The network observes how far the current control sequence misses the target and produces an improved sequence, capturing the refinement behavior of classical solvers in a form suitable for real-time embedded deployment.

% ==============================================================================
% PROPOSED APPROACH
% ==============================================================================

\section{Proposed Approach}
\label{sec:method}

The TRM framework demonstrates that weight sharing across refinement iterations yields dramatic parameter efficiency for discrete reasoning tasks~\cite{jolicoeurmartineau2024lessmore}. TRM applies the same reasoning blocks repeatedly, learning a refinement procedure rather than memorizing input-output mappings. For iterative problems, this principle yields order-of-magnitude efficiency gains.

TRC adapts this principle to continuous control by addressing two structural differences between discrete reasoning and trajectory optimization. First, discrete reasoning operates on tokens while control synthesis operates on continuous trajectories. Second, TRM refines textual answers through attention alone while TRC refines control sequences through physical simulation. The connection lies in the iterative refinement structure described in Section~\ref{sec:problem}: both TRM and classical trajectory optimization follow the same pattern of iterative improvement with a fixed update rule. Following TRM, TRC employs a two-level hierarchical latent structure: a high-level latent $\mathbf{z}_H$ that maintains strategic context and trajectory coordination, and a low-level latent $\mathbf{z}_L$ that handles tactical execution and immediate error correction. The same reasoning module processes both levels, achieving parameter efficiency through weight sharing across the hierarchy. The result is a compact network that achieves near-optimal performance while requiring only millisecond-scale inference.

This design offers three practical advantages. First, computational cost becomes adjustable: more iterations yield better solutions when time permits, while fewer iterations provide faster approximate solutions when latency is critical. Second, memory remains constant regardless of iteration count, since the same weights are reused. Third, the intermediate solutions $\mathbf{u}^{(1)}, \ldots, \mathbf{u}^{(K-1)}$ provide interpretability, allowing inspection of how the network refines its initial guess toward the final answer.

\subsection{Architecture}

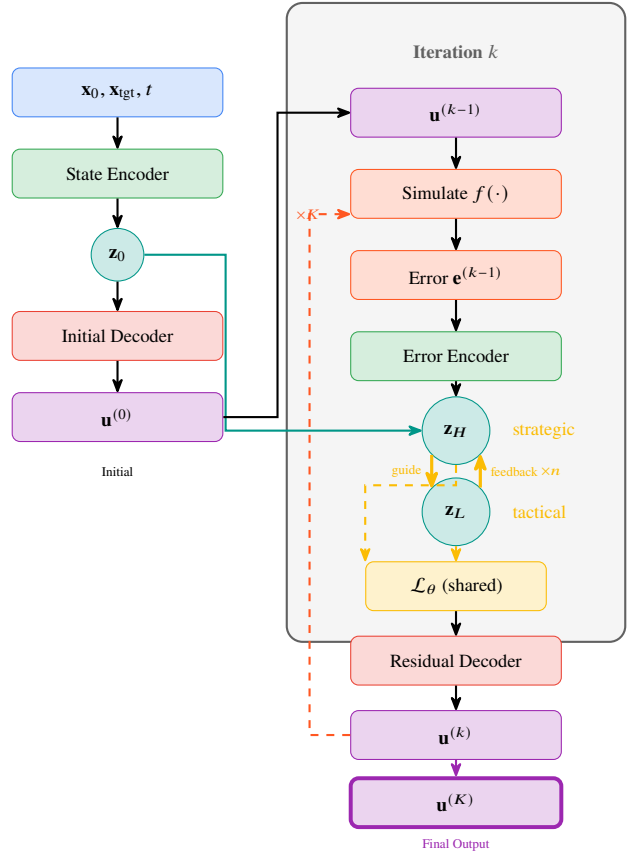
\begin{figure}[t]
\centering
\begin{tikzpicture}[
    scale=0.9,
    >={Stealth[round]},
    box/.style={rectangle, rounded corners=3pt, draw, minimum width=2.8cm, minimum height=0.65cm, align=center, font=\scriptsize},
    input/.style={box, fill=inputcolor!20, draw=inputcolor},
    encoder/.style={box, fill=encodercolor!20, draw=encodercolor},
    reasoning/.style={box, fill=reasoningcolor!20, draw=reasoningcolor},
    decoder/.style={box, fill=decodercolor!20, draw=decodercolor},
    output/.style={box, fill=outputcolor!20, draw=outputcolor},
    feedback/.style={box, fill=feedbackcolor!20, draw=feedbackcolor},
    latent/.style={circle, draw=latentcolor, fill=latentcolor!20, minimum size=0.7cm, font=\scriptsize},
    arrow/.style={->, thick},
]

% === LEFT COLUMN: Initial Generation ===
\node[input] (inputs) at (0, 4.5) {$\mathbf{x}_0, \mathbf{x}_{\text{tgt}}, t$};
\node[encoder] (enc) at (0, 3.3) {State Encoder};
\node[latent] (z0) at (0, 2.1) {$\mathbf{z}_0$};
\node[decoder] (init) at (0, 0.9) {Initial Decoder};
\node[output] (u0) at (0, -0.3) {$\mathbf{u}^{(0)}$};

\draw[arrow] (inputs) -- (enc);
\draw[arrow] (enc) -- (z0);
\draw[arrow] (z0) -- (init);
\draw[arrow] (init) -- (u0);

% === RIGHT COLUMN: One Refinement Iteration ===
% Box around refinement
\begin{scope}[on background layer]
\node[draw=darkgray, thick, rounded corners=6pt, fill=gray!8,
      minimum width=4.5cm, minimum height=8.5cm] at (5, 1.1) {};
\end{scope}
\node[font=\scriptsize\bfseries, text=darkgray] at (5, 5.1) {Iteration $k$};

\node[output] (uk_in) at (5, 4.2) {$\mathbf{u}^{(k-1)}$};
\node[feedback] (sim) at (5, 3.0) {Simulate $f(\cdot)$};
\node[feedback] (err) at (5, 1.8) {Error $\mathbf{e}^{(k-1)}$};
\node[encoder] (err_enc) at (5, 0.6) {Error Encoder};

% Two-level latent structure (increased size for visibility)
\node[latent, minimum size=0.9cm] (zH) at (5, -0.5) {$\mathbf{z}_H$};
\node[latent, minimum size=0.9cm] (zL) at (5, -1.7) {$\mathbf{z}_L$};
\node[font=\scriptsize, text=reasoningcolor, anchor=west] at (5.7, -0.5) {strategic};
\node[font=\scriptsize, text=reasoningcolor, anchor=west] at (5.7, -1.7) {tactical};

% n× cycle indicator
\node[font=\tiny, text=reasoningcolor] at (6.4, -1.1) {$\times n$};

% Bidirectional arrows between z_H and z_L (improved visibility)
\draw[arrow, reasoningcolor, line width=1.2pt] (zH.south west) -- (zL.north west) node[midway, left, font=\tiny] {guide};
\draw[arrow, reasoningcolor, line width=1.2pt] (zL.north east) -- (zH.south east) node[midway, right, font=\tiny] {feedback};

% Shared reasoning module (connected to both levels)
\node[reasoning, minimum width=2.4cm] (Ltheta) at (5, -2.8) {$\mathcal{L}_\theta$ (shared)};

% Arrows showing L_θ serves both z_H and z_L
\draw[arrow, reasoningcolor, dashed] (zH.south) -- ++(0, -0.3) -| (Ltheta.north west);
\draw[arrow, reasoningcolor, dashed] (zL.south) -- (Ltheta.north);

\node[decoder] (res) at (5, -3.9) {Residual Decoder};
\node[output] (uk_out) at (5, -5.0) {$\mathbf{u}^{(k)}$};

\draw[arrow] (uk_in) -- (sim);
\draw[arrow] (sim) -- (err);
\draw[arrow] (err) -- (err_enc);
\draw[arrow] (err_enc) -- (zH);
\draw[arrow] (Ltheta) -- (res);
\draw[arrow] (res) -- (uk_out);

% z0 input to z_H (from left)
\draw[arrow, latentcolor] (z0.east) -- ++(1.2, 0) |- (zH.west);

% === ARROWS BETWEEN COLUMNS ===
% u0 feeds first iteration
\draw[arrow] (u0.east) -- ++(0.8, 0) |- (uk_in.west);

% Loop back (dashed) - repeat K times
\draw[arrow, feedbackcolor, thick, dashed]
    (uk_out.west) -- ++(-0.6, 0) -- ++(0, 7.7) -- ++(0.6, 0)
    node[pos=0.5, left, font=\tiny, text=feedbackcolor] {$\times K$};

% Final output
\node[output, draw=outputcolor, line width=1.5pt] (final) at (5, -6.0) {$\mathbf{u}^{(K)}$};
\draw[arrow, outputcolor, thick] (uk_out) -- (final);

% Labels
\node[font=\tiny, anchor=north] at (0, -0.9) {Initial};
\node[font=\tiny, text=outputcolor, anchor=north] at (5, -6.4) {Final Output};

\end{tikzpicture}
\caption{TRC architecture. Left: initial control generation from encoded state. Right: one refinement iteration showing trajectory simulation, error computation, and the two-level latent structure. The high-level latent $\mathbf{z}_H$ maintains strategic context and guides the low-level latent $\mathbf{z}_L$, which handles tactical adjustments through $n$ cycles before feeding back to update $\mathbf{z}_H$. Both levels share the same reasoning module $\mathcal{L}_\theta$. The iteration repeats $K$ times with $\mathbf{z}_0$ conditioning each step.}
\label{fig:architecture}
\end{figure}

Figure~\ref{fig:architecture} illustrates the TRC architecture, which operates in two phases. In the initial generation phase, given the initial state $\mathbf{x}_0$, target state $\mathbf{x}_{\text{target}}$, and time horizon $T$, a state encoder produces a latent representation $\mathbf{z}_0$ and an initial decoder generates the first control estimate $\mathbf{u}^{(0)}$ directly from this encoding. This provides a reasonable starting point for refinement.

% Architecture Figure - Clean vertical flow design

In the iterative refinement phase, each iteration $k = 1, \ldots, K$ follows the same sequence: simulate the dynamics $f(\cdot)$ with current controls to obtain the predicted trajectory, compute the terminal tracking error $\mathbf{e}^{(k-1)} = \mathbf{x}_T^{(k-1)} - \mathbf{x}_{\text{target}}$, encode the error and current controls into latent representations, pass the combined encoding through $L$ reasoning blocks for $n$ inner cycles, and decode a control residual $\Delta\mathbf{u}^{(k)}$ to update the controls. The refinement operator can be written compactly as:
\begin{equation}
\label{eq:refinement}
\mathbf{u}^{(k)} = \mathcal{R}_\theta(\mathbf{u}^{(k-1)}, \mathbf{x}_0, \mathbf{x}_{\text{target}}, \mathbf{e}^{(k-1)})
\end{equation}
The critical design choice is that $\mathcal{R}_\theta$ uses identical parameters $\theta$ at every iteration. The reasoning blocks, which constitute the bulk of model parameters, are shared across all $K \times n$ applications. This weight sharing enables a 1.5M-parameter network to achieve near-optimal performance through iteration depth rather than parameter count.

The architecture comprises five neural network modules. The state encoder maps the control problem specification to a latent representation:
\begin{equation}
\mathbf{z}_0 = \text{MLP}_{\text{state}}([\mathbf{x}_0; \mathbf{x}_{\text{target}}; t_{\text{remaining}}])
\end{equation}
where $[\cdot;\cdot]$ denotes concatenation and $t_{\text{remaining}}$ encodes the time horizon. The encoder uses two linear layers with LayerNorm and GELU activation, mapping from $(2 d_x + 1)$ to $d_z$. This encoding $\mathbf{z}_0$ persists across all refinement iterations, providing consistent problem context.

The error encoder embeds trajectory tracking errors into the latent space:
\begin{equation}
\mathbf{z}_{\text{err}} = \text{MLP}_{\text{error}}(\mathbf{e}^{(k-1)})
\end{equation}
This feedback signal tells the network how far the current solution is from the goal. The control embedding projects the current control sequence to latent space via $\mathbf{z}_{\text{ctrl}} = \text{Linear}(\text{flatten}(\mathbf{u}^{(k-1)}))$, allowing the reasoning module to condition on the controls being refined.

The reasoning module forms the core of TRC, employing a two-level hierarchical structure inspired by TRM. Two separate latent states capture different aspects of the control problem: the high-level latent $\mathbf{z}_H$ encodes strategic planning and overall trajectory coordination, while the low-level latent $\mathbf{z}_L$ encodes tactical execution and detailed control adjustments. Both states are initialized by combining learnable parameters $\mathbf{H}_{\text{init}}$ and $\mathbf{L}_{\text{init}}$ with sample-specific projections from the problem encoding $\mathbf{z}_0$, ensuring each sample begins with distinct latent representations.

At each outer iteration $k$, the context encoding combines problem specification with feedback:
\begin{equation}
\mathbf{z}_{\text{ctx}} = \mathbf{z}_0 + \mathbf{z}_{\text{err}} + \mathbf{z}_{\text{ctrl}}
\end{equation}
The low-level latent then processes this context through $n$ tactical cycles, receiving guidance from the high-level state:
\begin{equation}
\mathbf{z}_L^{(i+1)} = \mathcal{L}_\theta(\mathbf{z}_L^{(i)}, \mathbf{z}_H + \mathbf{z}_{\text{ctx}})
\end{equation}
After the tactical cycles complete, the high-level latent updates once based on what the low-level learned:
\begin{equation}
\mathbf{z}_H^{(k+1)} = \mathcal{L}_\theta(\mathbf{z}_H^{(k)}, \mathbf{z}_L^{(n)})
\end{equation}
The critical insight from TRM is that the same reasoning module $\mathcal{L}_\theta$ processes both levels. This weight sharing achieves parameter efficiency: rather than separate networks for strategy and tactics, a single module learns to reason at both levels depending on its input context. The module applies multi-head self-attention and feed-forward layers with residual connections. The attention mechanism enables dynamic weighting between problem context, error feedback, and control history.

Two decoders generate control outputs. The initial decoder produces the first control estimate from the state encoding via $\mathbf{u}^{(0)} = \text{MLP}_{\text{initial}}(\mathbf{z}_0)$. The residual decoder generates control corrections during refinement:
\begin{equation}
\Delta\mathbf{u}^{(k)} = \text{MLP}_{\text{residual}}([\mathbf{z}_H^{(k)}; \mathbf{u}^{(k-1)}])
\end{equation}
The network predicts control corrections rather than regenerating the full sequence from scratch because corrections are typically small relative to control magnitudes (making them easier to learn) and residual updates preserve temporal smoothness in the control sequence. Controls are updated via residual addition with clipping to enforce bounds:
\begin{equation}
\mathbf{u}^{(k)} = \text{clip}(\mathbf{u}^{(k-1)} + \Delta\mathbf{u}^{(k)}, u_{\min}, u_{\max})
\end{equation}

The complete TRC forward pass integrates the five components described above---state encoder, error encoder, two-level reasoning module, initial decoder, and residual decoder---into a coherent refinement loop. Algorithm~\ref{alg:trc} details this integration: given an initial state and target, TRC first encodes the problem and generates an initial control estimate, then iteratively simulates, observes error, reasons at both levels, and applies corrections.

\begin{algorithm}[htb!]
\caption{TRC Two-Level Recursive Control Synthesis}
\label{alg:trc}
\begin{algorithmic}[1]
\REQUIRE Initial state $\mathbf{x}_0$, target $\mathbf{x}_{\text{target}}$, dynamics $f(\cdot)$
\ENSURE Refined control sequence $\mathbf{u}^{(K)}$
\STATE $\mathbf{z}_0 \gets \text{StateEncoder}(\mathbf{x}_0, \mathbf{x}_{\text{target}}, t_{\text{remaining}})$
\STATE $\mathbf{z}_H \gets \mathbf{H}_{\text{init}}$ \COMMENT{Learnable strategic initialization}
\STATE $\mathbf{z}_L \gets \mathbf{L}_{\text{init}}$ \COMMENT{Learnable tactical initialization}
\STATE $\mathbf{u}^{(0)} \gets \text{InitialDecoder}(\mathbf{z}_0)$
\FOR{$k = 1$ to $K$}
    \STATE $\mathbf{x}_{1:T}^{(k-1)} \gets \text{Simulate}(f, \mathbf{x}_0, \mathbf{u}^{(k-1)})$
    \STATE $\mathbf{e}^{(k-1)} \gets \mathbf{x}_T^{(k-1)} - \mathbf{x}_{\text{target}}$
    \STATE $\mathbf{z}_{\text{ctx}} \gets \mathbf{z}_0 + \text{ErrorEmbed}(\mathbf{e}^{(k-1)}) + \text{ControlEmbed}(\mathbf{u}^{(k-1)})$
    \FOR{$i = 1$ to $n$} % Low-level tactical cycles
        \STATE $\mathbf{z}_L \gets \mathcal{L}_\theta(\mathbf{z}_L, \mathbf{z}_H^{(k-1)} + \mathbf{z}_{\text{ctx}})$
    \ENDFOR
    \STATE $\mathbf{z}_H^{(k)} \gets \mathcal{L}_\theta(\mathbf{z}_H^{(k-1)}, \mathbf{z}_L)$ \COMMENT{High-level strategic update}
    \STATE $\Delta\mathbf{u}^{(k)} \gets \text{ResidualDecoder}(\mathbf{z}_H^{(k)}, \mathbf{u}^{(k-1)})$
    \STATE $\mathbf{u}^{(k)} \gets \text{clip}(\mathbf{u}^{(k-1)} + \Delta\mathbf{u}^{(k)}, u_{\min}, u_{\max})$
\ENDFOR
\RETURN $\mathbf{u}^{(K)}$
\end{algorithmic}
\end{algorithm}

Model size scales with problem complexity. For the Van der Pol and powered descent problems ($d_z = 256$, $d_h = 512$, $L = 3$ blocks, 8 heads), the model uses approximately 1.5M parameters. The shared reasoning module is applied $K \times (n + 1)$ times per forward pass: with $K=3$ outer iterations and $n=4$--6 inner cycles, this means 15--21 applications of the same module, yielding effective computational depth of 15--21 sequential module applications while maintaining constant memory (under 10 MB for weights). Beyond memory efficiency, weight sharing imposes a structural constraint: the network must learn a general refinement operator that improves any control sequence, not iteration-specific corrections that only work at particular stages of convergence. Compared to LLM approaches with 50M+ trainable parameters, TRC achieves a 95--99\% reduction.

\subsection{Theoretical Foundation: Learned Gradient Descent}

The TRC refinement operator can be interpreted as learned gradient descent on the trajectory cost, a perspective that provides insight into why the architecture succeeds and suggests connections to classical optimization theory.

Consider the terminal cost as a function of the control sequence:
\begin{equation}
\label{eq:terminal_cost}
J(\mathbf{u}) = \frac{1}{2}\|\mathbf{x}_T(\mathbf{u}) - \mathbf{x}_{\text{target}}\|_2^2 = \frac{1}{2}\|\mathbf{e}(\mathbf{u})\|_2^2
\end{equation}
where the terminal state $\mathbf{x}_T(\mathbf{u})$ depends on the control sequence through the dynamics $\mathbf{x}_{t+1} = f(\mathbf{x}_t, \mathbf{u}_t)$ for $t = 0, \ldots, T-1$. The goal of refinement is to find controls $\mathbf{u}^*$ that minimize this cost. Applying the chain rule, the gradient of $J$ with respect to the control sequence is:
\begin{equation}
\label{eq:gradient}
\nabla_{\mathbf{u}} J = \left(\frac{\partial \mathbf{x}_T}{\partial \mathbf{u}}\right)^\top (\mathbf{x}_T - \mathbf{x}_{\text{target}}) = \left(\frac{\partial \mathbf{x}_T}{\partial \mathbf{u}}\right)^\top \mathbf{e}
\end{equation}
The Jacobian $\partial \mathbf{x}_T / \partial \mathbf{u} \in \mathbb{R}^{d_x \times (T \cdot d_u)}$ captures how the terminal state responds to control perturbations, depending on the system dynamics and the current trajectory.

Standard gradient descent updates controls as $\mathbf{u}^{(k)} = \mathbf{u}^{(k-1)} - \eta \nabla_{\mathbf{u}} J(\mathbf{u}^{(k-1)})$. Comparing with the TRC update $\mathbf{u}^{(k)} = \mathbf{u}^{(k-1)} + \Delta\mathbf{u}^{(k)}$, the high-level latent $\mathbf{z}_H$ learns to encode the strategic descent direction:
\begin{equation}
\label{eq:learned_gradient}
\Delta\mathbf{u}^{(k)} = \text{ResidualDecoder}(\mathbf{z}_H^{(k)}, \mathbf{u}^{(k-1)}) \approx -\eta \nabla_{\mathbf{u}} J(\mathbf{u}^{(k-1)})
\end{equation}
The error vector $\mathbf{e}^{(k-1)}$ provides the raw gradient direction information per Eq.~\eqref{eq:gradient}, while the error encoder extracts features relevant to computing the full gradient. Within each outer iteration, the low-level latent $\mathbf{z}_L$ refines tactical adjustments through $n$ cycles of processing, accumulating local corrections before passing this information to $\mathbf{z}_H$. The high-level latent then integrates these tactical refinements into a coherent strategic update. Through this two-level process, the shared reasoning module $\mathcal{L}_\theta$ learns to approximate the sensitivity matrix $\partial \mathbf{x}_T / \partial \mathbf{u}$ implicitly through training on optimal trajectories. This learned approach is more expressive than classical gradient descent in two ways: the network can learn state-dependent step sizes rather than using a fixed $\eta$ (taking smaller steps in regions of high curvature and larger steps in flat regions), and through the reasoning module, the network can incorporate curvature information to approximate Newton-like updates without explicitly computing Hessians.

Three observations from experiments support the gradient descent interpretation. Trajectory error decreases monotonically across iterations in the majority of test cases, consistent with gradient descent on a smooth cost landscape. The magnitude $\|\Delta\mathbf{u}^{(k)}\|$ decreases with iteration count $k$, analogous to approaching a minimum where gradients vanish. Control updates consistently point in directions that reduce terminal error, confirming that the learned residuals approximate descent directions.

The gradient descent perspective also explains why the same network can apply at every iteration: the refinement operation is fundamentally the same (compute gradient, take step) regardless of iteration count. Just as gradient descent uses the same update rule at every step, TRC uses the same refinement operator $\mathcal{R}_\theta$. The network learns a procedure for improvement rather than a mapping to the answer.

% ==============================================================================
% TRAINING METHODOLOGY (Section 4)
% ==============================================================================

\section{Training Methodology}
\label{sec:training}

Standard behavior cloning trains a policy $\pi_\theta(\mathbf{x}) \to \mathbf{u}$ to directly predict optimal controls using $\mathcal{L}_{\text{BC}} = \|\pi_\theta(\mathbf{x}) - \mathbf{u}^*\|_2^2$, supervising only the final output and providing no learning signal for the refinement process itself. TRC instead uses process supervision: in addition to matching the final controls, the network is rewarded for improving trajectory cost at each iteration. The loss function is:
\begin{equation}
\label{eq:loss}
\mathcal{L} = \underbrace{\|\mathbf{u}^{(K)} - \mathbf{u}^*\|_2^2}_{\text{final accuracy}} - \lambda \underbrace{\frac{1}{K-1}\sum_{k=1}^{K-1} \left( \tilde{J}^{(k-1)} - \tilde{J}^{(k)} \right)}_{\text{improvement reward}}
\end{equation}
where $\tilde{J}^{(k)} = J^{(k)} / J^{(0)}$ is the normalized trajectory cost after iteration $k$ (normalized by the initial cost to ensure scale-invariance), and $\lambda$ controls the strength of process supervision (set to $\lambda = 0.1$ to $0.5$ depending on the problem). The improvement reward encourages cost reduction across iterations: the network receives higher reward when each refinement step decreases trajectory cost. This teaches the refinement operator to improve at each step, not merely produce the correct final answer through arbitrary intermediate computations. Without process supervision, the network could learn to ignore early iterations entirely, treating the full $K$-iteration stack as a single complex function. Process supervision ensures each iteration contributes meaningful refinement, which is essential for the property that more iterations yield better results. This approach draws inspiration from process supervision in language model reasoning~\cite{lightman2023let}, adapted here for continuous control where reasoning steps correspond to refinement iterations.

During training, the \textit{improvement metric} quantifies the average benefit per refinement iteration:
\begin{equation}
\label{eq:improvement}
\text{Improvement} = \frac{1}{B(K-1)} \sum_{b=1}^{B} \sum_{k=0}^{K-2} \frac{J_b^{(k)} - J_b^{(k+1)}}{J_b^{(0)}}
\end{equation}
where $B$ is the batch size, $J_b^{(k)}$ is the trajectory cost for sample $b$ after iteration $k$, and costs are normalized by the initial cost $J_b^{(0)}$. This metric averages the relative cost reduction across all refinement transitions and batch samples. A value of 0.25 indicates that each iteration reduces cost by 25\% of the initial value on average; with $K=3$ iterations (two transitions), the total relative improvement would be approximately $0.25 \times 2 = 50\%$.

Training requires access to optimal control sequences, which serve as supervision targets. The specific datasets and their generation methods are described in Section~\ref{sec:experiments} for each problem. TRC trains on modest datasets (thousands of samples), three orders of magnitude fewer than typical deep learning applications. This efficiency stems from the recursive architecture: rather than learning a distinct mapping for each initial condition, the network learns a single refinement operator that generalizes across conditions. The supervision signal is also richer since each training example provides $K+1$ targets (one per iteration), effectively multiplying the dataset size. Language models for control~\cite{brohan2023rt2} require millions of demonstrations and billions of parameters; TRC achieves near-optimal control costs with 99\% fewer parameters.

Training uses the AdamW optimizer with learning rate $1 \times 10^{-3}$, batch sizes of 32--64, and cosine annealing schedule. Van der Pol requires 50 epochs (approximately 30 minutes on a single NVIDIA RTX 3080 GPU), while powered descent requires 200 epochs (approximately 2 hours) due to the higher-dimensional state space. Gradient clipping with max norm 1.0 ensures stability when backpropagating through the dynamics simulation within the refinement loop.

% ==============================================================================
% NUMERICAL RESULTS (Section 5)
% ==============================================================================

\section{Numerical Results}
\label{sec:experiments}

TRC is evaluated on two nonlinear control problems of increasing complexity, progressing from low-dimensional to high-dimensional state spaces. Table~\ref{tab:problems} summarizes the problem specifications. The same TRC architecture handles both problems, with $K=3$ refinement iterations and $n=4$--6 inner cycles. Dataset details and optimal solution methods are described in each subsection.

\begin{table}[h]
\centering
\caption{Control Problem Specifications}
\label{tab:problems}
\begin{tabular}{lcccc}
\hline
\textbf{Problem} & \textbf{State Dim} & \textbf{Control Dim} & \textbf{Dynamics} & \textbf{Horizon} \\
\hline
Van der Pol & 2 & 1 & Nonlinear & 100 \\
Powered Descent & 7 & 3 & Nonlinear, variable-mass & 50 \\
\hline
\end{tabular}
\end{table}

\subsection{Van der Pol Oscillator}

\begin{figure}[h]
\centering
\includegraphics[width=0.6\columnwidth]{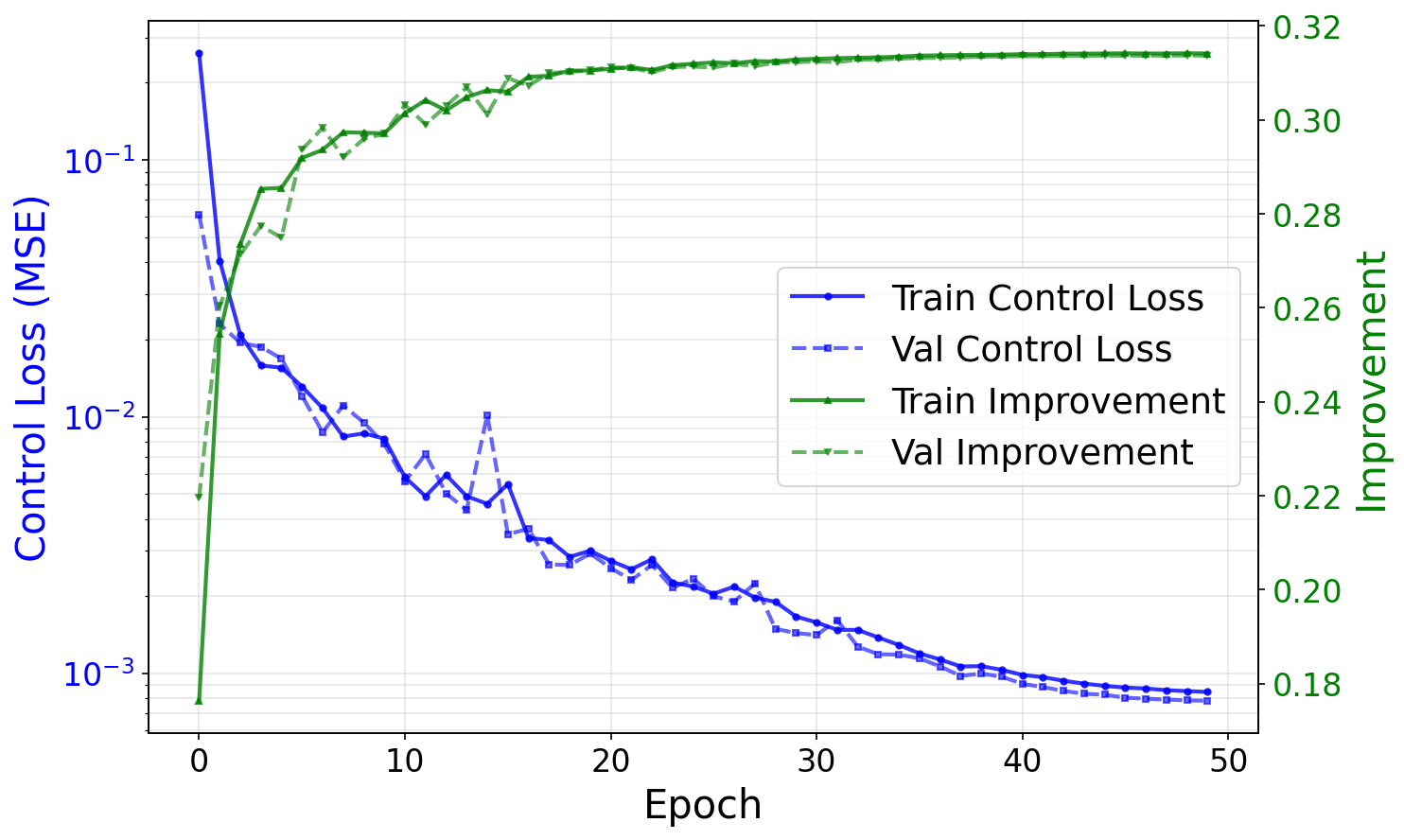}
\caption{Van der Pol oscillator training convergence: control loss (blue) and improvement metric (green) over 50 epochs.}
\label{fig:vdp_training}
\end{figure}

\begin{figure}[hb!]
\centering
\includegraphics[width=0.9\columnwidth]{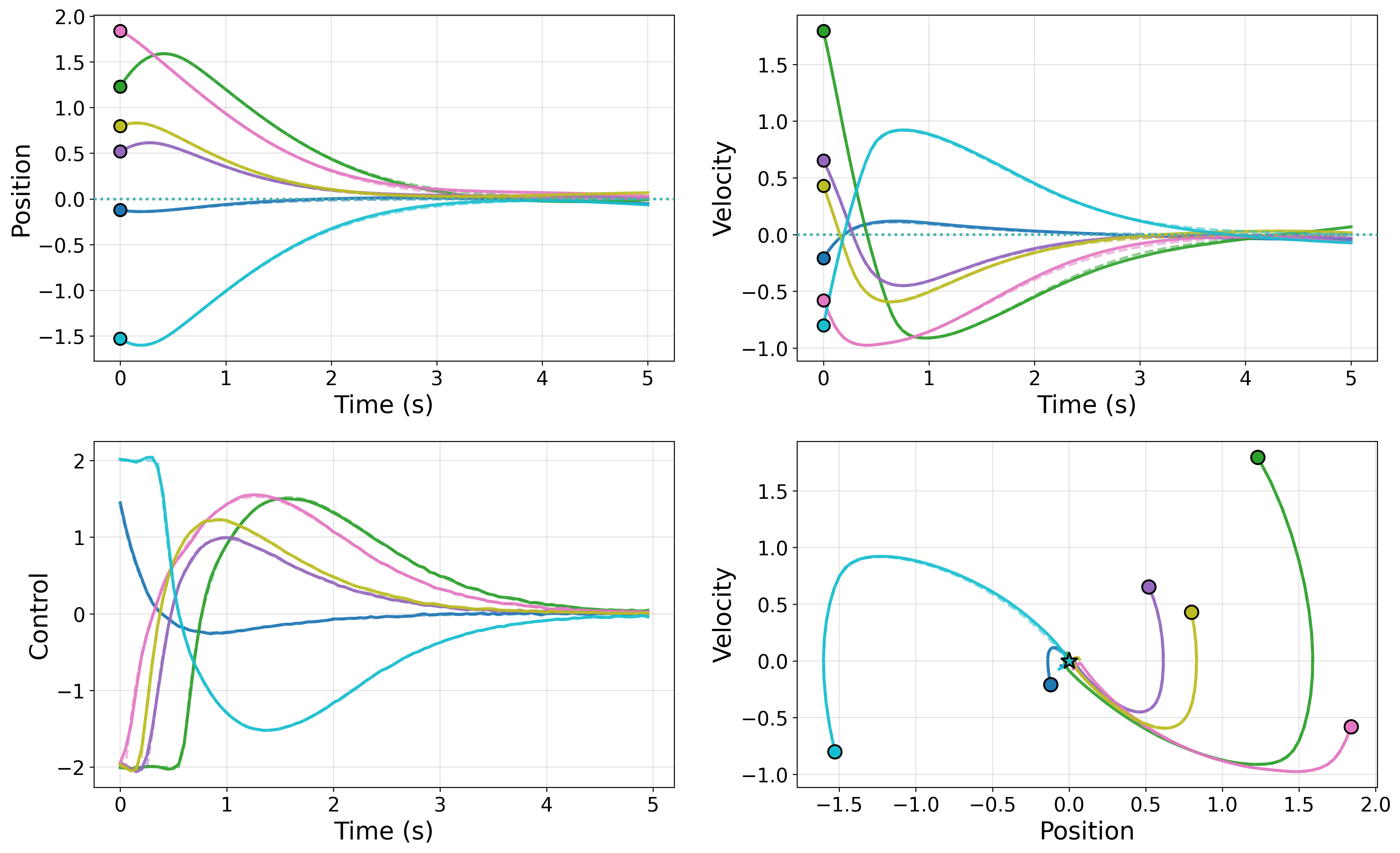}
\caption{Van der Pol oscillator trajectory results. All trajectories converge to the origin (target state), with the phase portrait showing successful stabilization from diverse initial conditions in the nonlinear regime.}
\label{fig:vdp_traj}
\end{figure}

The Van der Pol oscillator is a classical nonlinear system exhibiting self-sustained oscillations, originally developed to model vacuum tube circuits and now widely used as a benchmark for nonlinear control. The dynamics are:
\begin{equation}
\label{eq:vanderpol}
\ddot{x} - \mu(1 - x^2)\dot{x} + x = u
\end{equation}
In state-space form with $\mathbf{x} = [x, \dot{x}]^\top$:
\begin{equation}
\dot{\mathbf{x}} = \begin{bmatrix} x_2 \\ \mu(1 - x_1^2)x_2 - x_1 + u \end{bmatrix}
\end{equation}
The parameter $\mu = 1.0$ places the system in the weakly nonlinear regime, where the uncontrolled system exhibits a stable limit cycle. The state-dependent damping term $\mu(1 - x^2)\dot{x}$ provides negative damping when $|x| < 1$ (energy injection) and positive damping when $|x| > 1$ (energy dissipation), sustaining periodic oscillations. Stabilization to the origin requires overcoming this energy-pumping mechanism.

The system is integrated using fourth-order Runge-Kutta with time step $\Delta t = 0.05$ s over a horizon of $T = 100$ steps (5 seconds total). This extended horizon accommodates the oscillator's natural period of approximately $2\pi$ seconds. The control is bounded as $|u| \leq 2.0$. Optimal solutions are computed via Sequential Quadratic Programming (SQP) with cost function:
\begin{equation}
J = \sum_{t=0}^{T-1} \left( \mathbf{x}_t^\top \mathbf{Q} \mathbf{x}_t + R u_t^2 \right) + \mathbf{x}_T^\top \mathbf{Q}_f \mathbf{x}_T
\end{equation}
where $\mathbf{Q} = \text{diag}(10, 5)$, $R = 0.5$, and $\mathbf{Q}_f = 20 \cdot \mathbf{Q}$ to emphasize terminal accuracy. Initial states are sampled uniformly from $[-2, 2]^2$ with target state at the origin, yielding 10,000 training samples and 1,000 test samples.

Training dynamics are shown in Figure~\ref{fig:vdp_training}. The improvement metric (Eq.~\ref{eq:improvement}) reaches approximately 0.32, reflecting the benefit of iterative correction for nonlinear dynamics. Figure~\ref{fig:vdp_traj} shows trajectory results: the model successfully stabilizes the oscillator from various initial conditions, driving both position and velocity to zero. The phase portrait demonstrates effective nonlinear control synthesis, with TRC achieving mean control cost of 79.6, matching the optimal cost exactly.

\begin{figure}[htb!]
\centering
\includegraphics[width=\columnwidth]{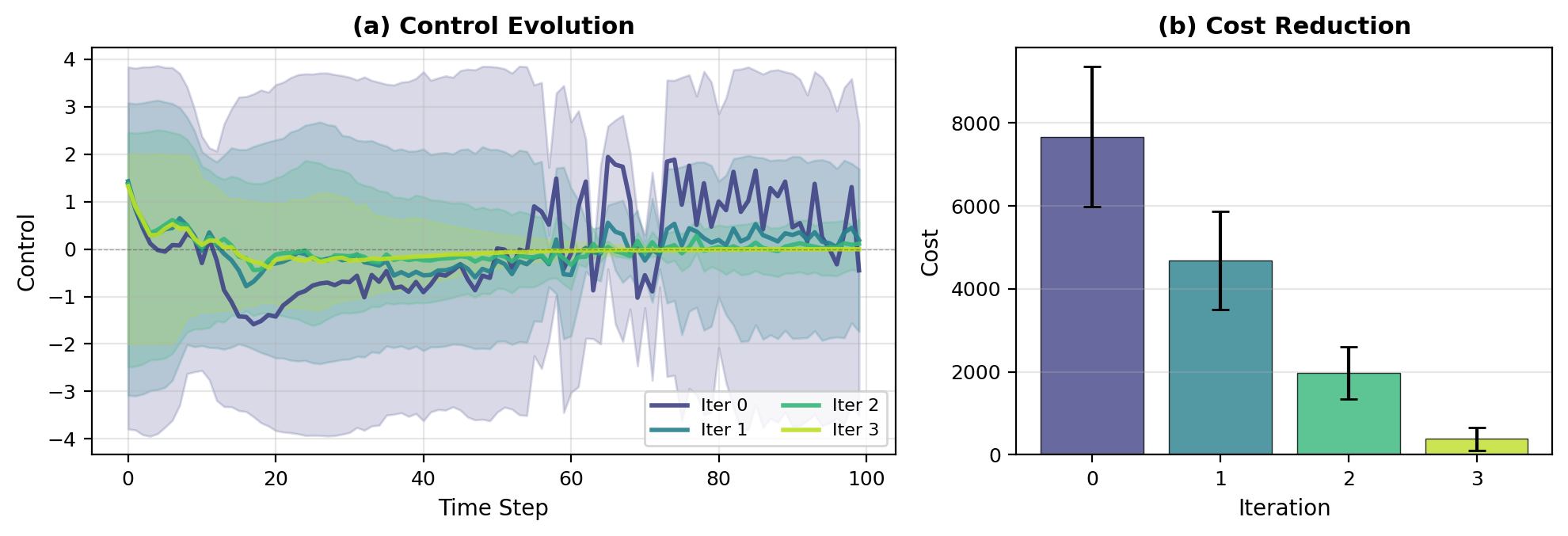}
\caption{Iterative refinement for Van der Pol oscillator. (a) Control evolution showing dramatically reduced variance across iterations. Initial predictions (purple) have high uncertainty; final iteration (yellow) converges to consistent damping strategy. (b) Cost reduction of approximately 90\% from iteration 0 to 3.}
\label{fig:vdp_refinement}
\end{figure}

\begin{figure}[htb!]
\centering
\includegraphics[width=\columnwidth]{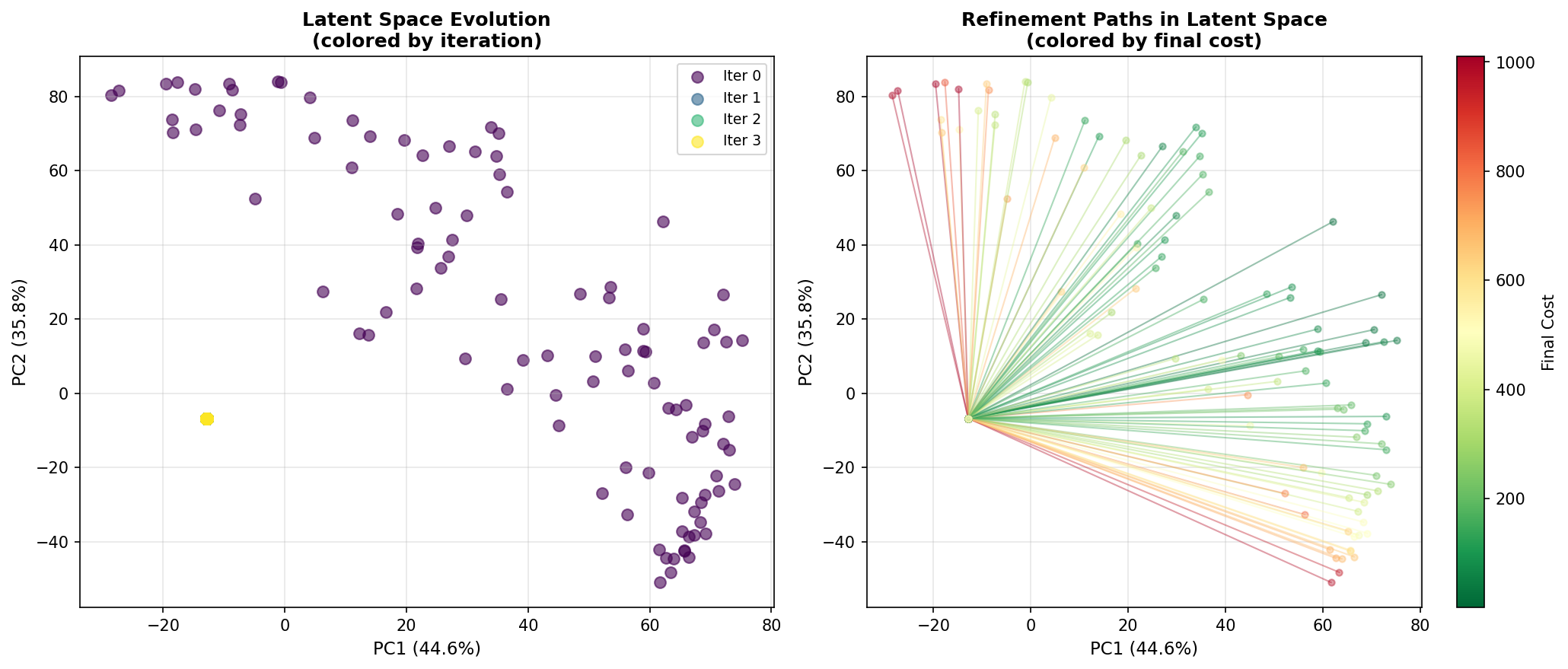}
\caption{Latent space evolution for Van der Pol oscillator. (Left) All samples converge from scattered initial states to a single attractor point, indicating the model discovers a universal control strategy. (Right) Refinement paths show consistent flow toward low-cost solutions regardless of initial condition.}
\label{fig:vdp_latent}
\end{figure}

The value of iterative refinement is particularly evident for nonlinear systems. Figure~\ref{fig:vdp_refinement} shows dramatic variance reduction across iterations: initial control predictions (purple) exhibit high uncertainty spanning $\pm 4$ control units, while the final iteration (yellow) converges to a consistent oscillation-damping strategy. Cost reduction exceeds 90\% from iteration 0 to 3. The latent space in Figure~\ref{fig:vdp_latent} exhibits a striking convergence phenomenon: scattered initial latent states (encoding diverse starting positions on the limit cycle) rapidly converge to a single attractor point. This indicates that TRC learns a universal damping strategy: the high-level latent $\mathbf{z}_H$ converges to a common representation while the residual decoder generates state-specific control corrections based on the current trajectory error.

\begin{figure}[h]
\centering
\includegraphics[width=0.6\columnwidth]{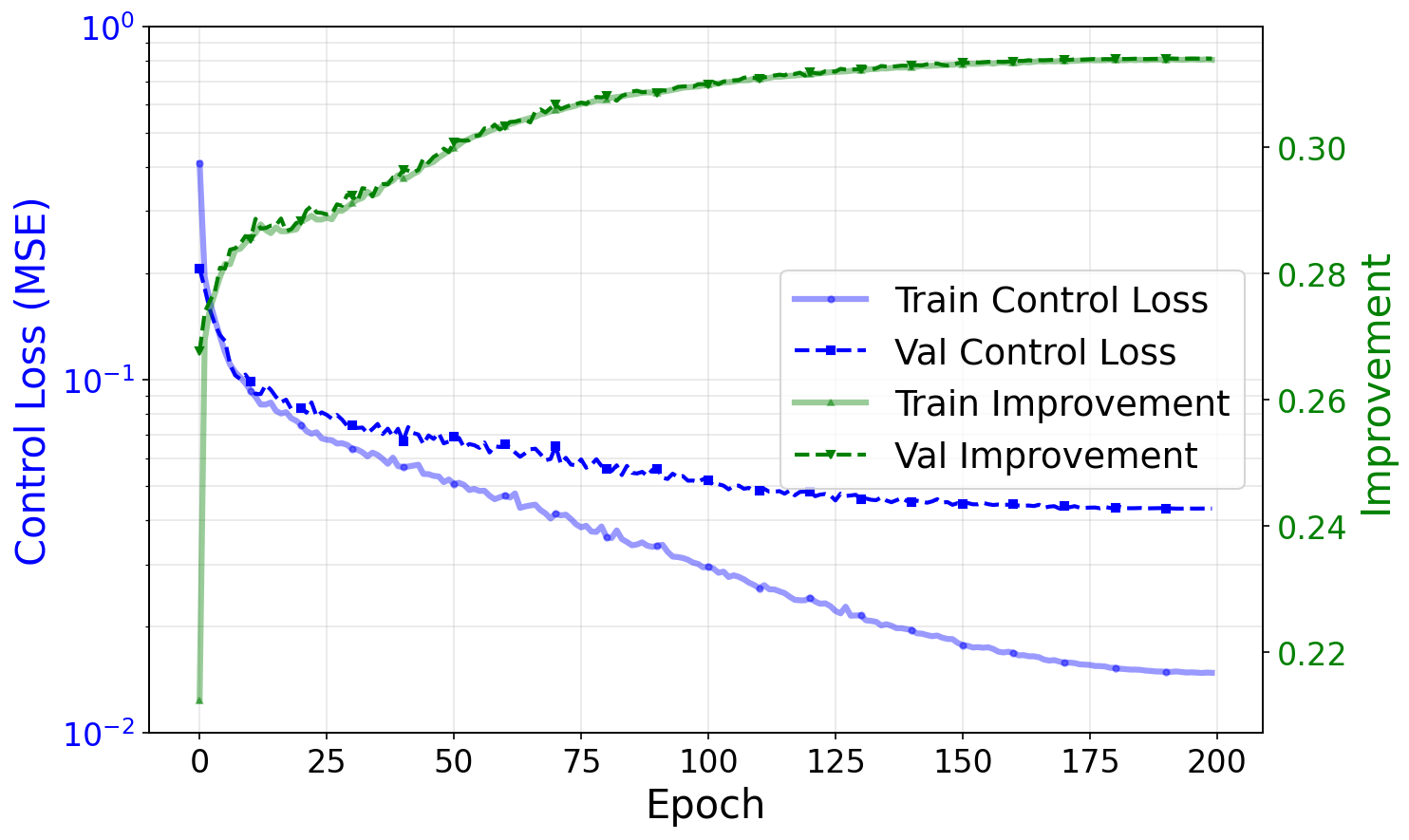}
\caption{Powered descent training convergence: control loss (blue) and improvement metric (green) over 200 epochs.}
\label{fig:rocket_training}
\end{figure}

\subsection{Powered Descent Guidance}

The powered descent guidance problem represents a critical phase of planetary landing, requiring fuel-optimal trajectory design subject to multiple constraints. This section summarizes the problem formulation; complete details including derivations, constraint handling, and the successive convexification solution method are provided by Briden et al.~\cite{briden2023tpdg}, which builds on the convex optimization framework of A\c{c}\i kme\c{s}e and Ploen~\cite{acikmese2007convex}. All problem parameters and the training dataset used in this work are taken directly from Briden et al.

The spacecraft state comprises position $\mathbf{r} = [x, y, z]^\top$ in a planet-fixed frame (with $z$ as altitude), velocity $\mathbf{v} = [\dot{x}, \dot{y}, \dot{z}]^\top$, and wet mass $m$. The equations of motion are:
\begin{equation}
\label{eq:rocket}
\dot{\mathbf{r}} = \mathbf{v}, \quad \dot{\mathbf{v}} = \frac{\mathbf{T}}{m} + \mathbf{g}, \quad \dot{m} = -\frac{\|\mathbf{T}\|_2}{I_{\text{sp}} g_0}
\end{equation}
where $\mathbf{T} \in \mathbb{R}^3$ is the thrust vector, $\mathbf{g} = [0, 0, -g_{\text{Mars}}]^\top$ with $g_{\text{Mars}} = 3.71$ m/s$^2$, $I_{\text{sp}} = 200.7$ s is the specific impulse, and $g_0 = 9.81$ m/s$^2$ is Earth's gravitational acceleration.

The optimization is subject to several constraints. Thrust bounds enforce engine operating limits:
\begin{equation}
T_{\min} \leq \|\mathbf{T}\|_2 \leq T_{\max}
\end{equation}
with $T_{\min} = 4000$ N and $T_{\max} = 13000$ N, reflecting throttle constraints typical of bipropellant engines. The glideslope constraint ensures terrain clearance during descent:
\begin{equation}
\|\mathbf{r}_{xy}\|_2 \leq z \tan(\gamma_{\text{gs}})
\end{equation}
where $\mathbf{r}_{xy} = [x, y]^\top$ is the horizontal position and $\gamma_{\text{gs}} = 75°$ is the glideslope angle measured from vertical. Terminal constraints require soft landing at the target:
\begin{equation}
\mathbf{r}(t_f) = \mathbf{0}, \quad \|\mathbf{v}(t_f)\|_2 \leq v_{\text{tol}}
\end{equation}
with $v_{\text{tol}} = 1.0$ m/s for safe touchdown. Mass bounds ensure feasibility:
\begin{equation}
m_{\text{dry}} \leq m(t) \leq m_{\text{wet}}
\end{equation}
with $m_{\text{dry}} = 1000$ kg and $m_{\text{wet}} = 2000$ kg.

The fuel-optimal objective minimizes propellant consumption, expressed in Mayer form as:
\begin{equation}
\label{eq:fuel_cost}
J = m_0 - m(t_f)
\end{equation}
where $m_0$ is the initial wet mass and $m(t_f)$ is the vehicle mass at touchdown. Maximizing final mass is equivalent to minimizing fuel consumed~\cite{briden2023tpdg}. The training dataset comprises 4,812 optimal trajectories from Briden et al., each generated via successive convexification solving this fuel-optimal problem subject to the dynamics~\eqref{eq:rocket} and constraints above.

The problem is discretized with variable time steps (mean $\Delta t \approx 1.15$ s) over $T = 50$ steps. Initial conditions are sampled from realistic powered descent initiation states: altitude $z_0 \in [1500, 2500]$ m, horizontal position $\|\mathbf{r}_{xy,0}\|_2 \leq 500$ m, downward velocity $\dot{z}_0 \in [-100, -50]$ m/s, horizontal velocity $\|\mathbf{v}_{xy,0}\|_2 \leq 50$ m/s, and initial mass $m_0 \in [1800, 2000]$ kg.

Figure~\ref{fig:rocket_training} shows training convergence over 200 epochs. The control loss (blue) decreases rapidly during the first 50 epochs as the network learns the coarse trajectory structure, then continues gradual refinement for the remaining epochs as it captures finer control details. This extended training duration compared to Van der Pol (200 vs.\ 50 epochs) reflects the increased complexity of the 7-dimensional state space and 3-dimensional control output. The improvement metric (green, Eq.~\ref{eq:improvement}) stabilizes around 0.32, remarkably matching the Van der Pol value despite the higher-dimensional problem. This consistency suggests that iterative refinement achieves similar relative cost reduction regardless of state-space dimension, with each iteration reducing trajectory cost by approximately one-third of the initial value.

\begin{figure}[htb!]
\centering
\hspace{-0.5in}
\begin{tabular}{cc}
\includegraphics[width=0.6\columnwidth]{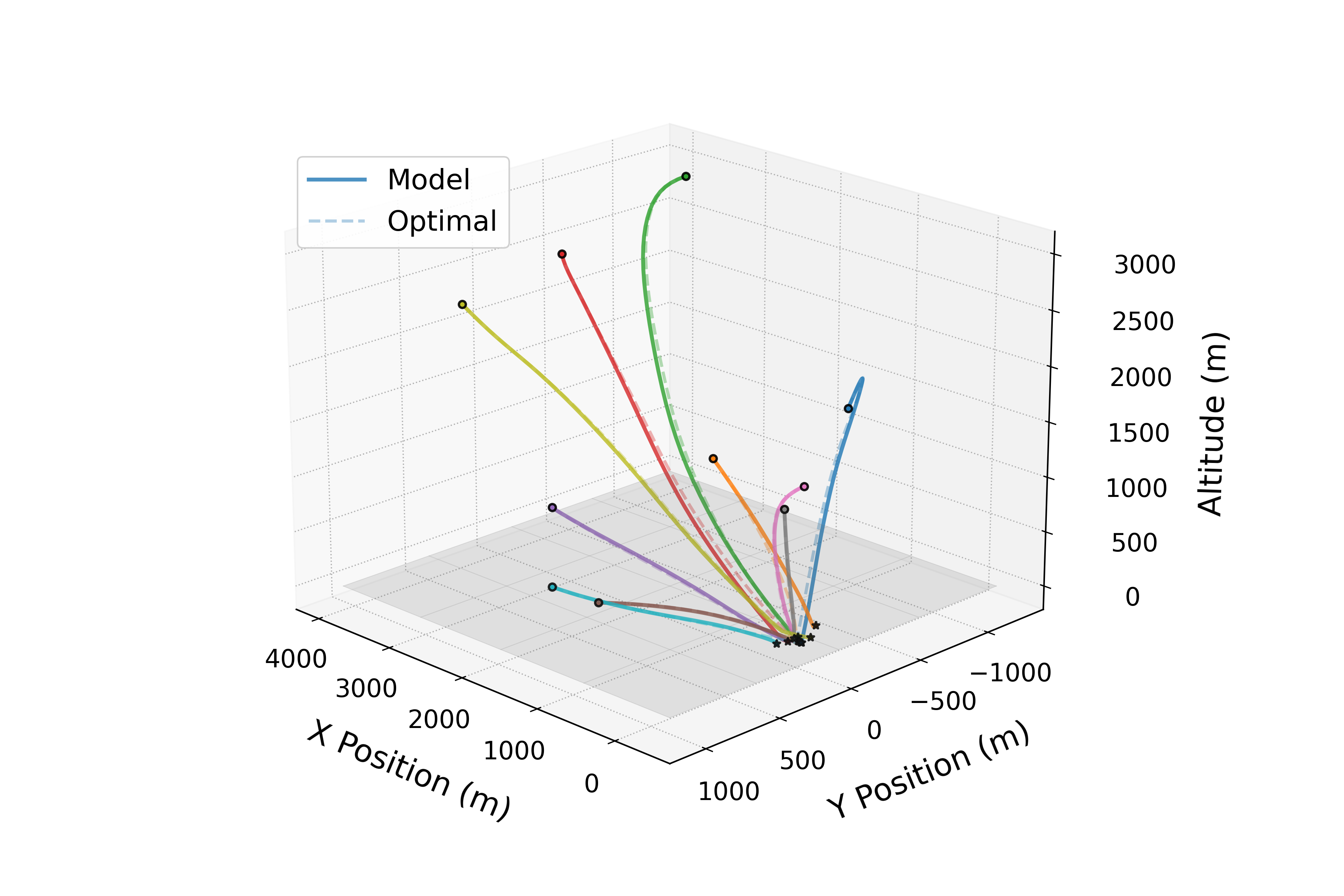} &
\includegraphics[width=0.4\columnwidth]{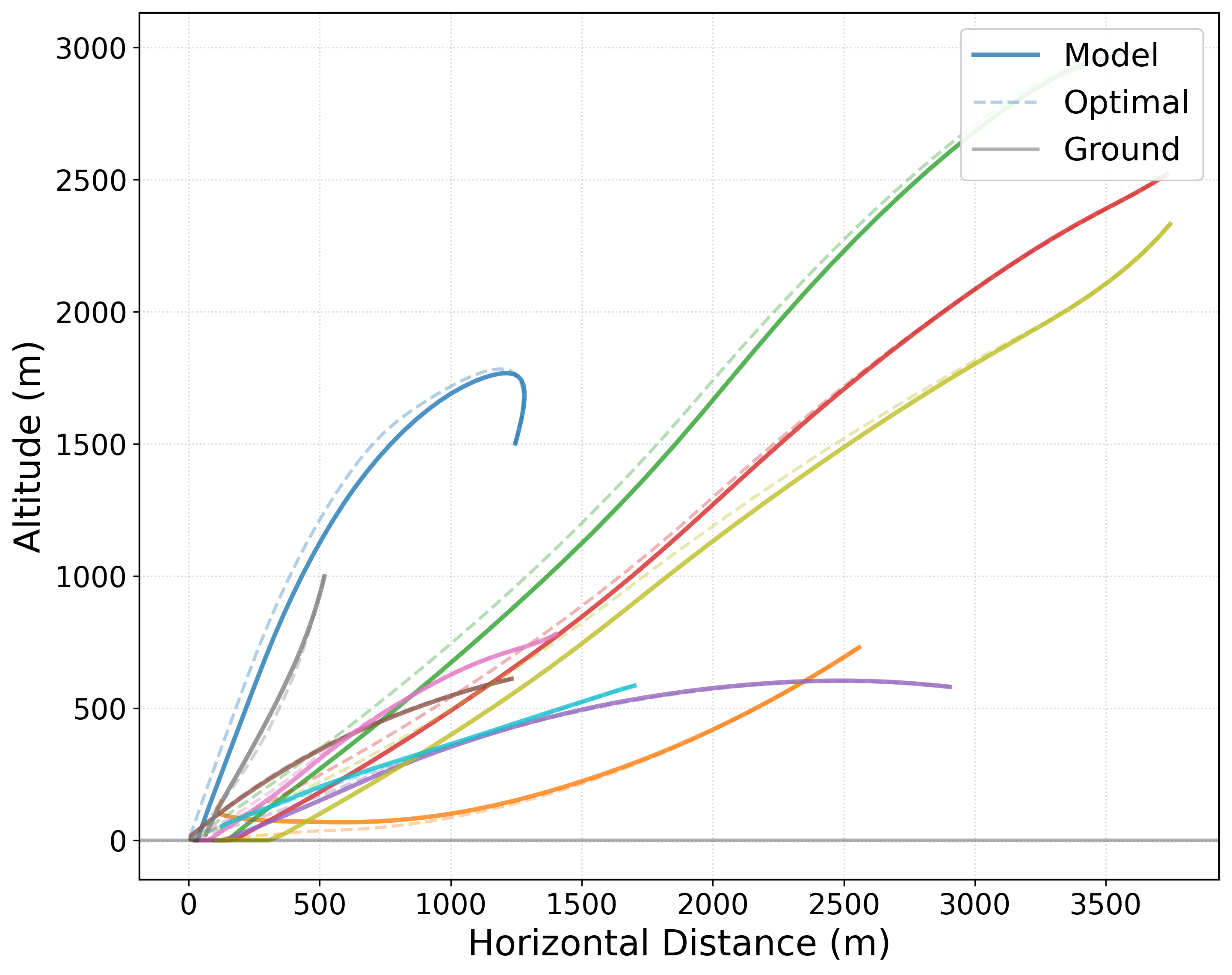}
\end{tabular}
\caption{Rocket landing trajectory results. Left: 3D view showing multiple descent trajectories from varied initial positions, all converging to the landing pad. Right: approach profile (altitude vs. horizontal distance) comparing TRC (solid) with optimal (dashed) solutions.}
\label{fig:rocket_traj}
\end{figure}

Figure~\ref{fig:rocket_traj} presents trajectory results from the test set. The 3D view (left) shows multiple descent trajectories originating from varied initial conditions---altitudes spanning 1500--2500~m and horizontal positions within a 500~m radius---all converging to the landing pad at the origin. The trajectories remain within the glideslope constraint cone ($\gamma_{\text{gs}} = 75°$), ensuring adequate terrain clearance throughout descent. The approach profile (right) compares TRC predictions (solid) with optimal solutions (dashed), demonstrating close agreement in altitude versus horizontal distance. The characteristic curved approach paths reflect fuel-optimal trajectory shaping: the vehicle trades horizontal velocity for vertical deceleration while respecting thrust constraints.

\begin{figure}[htb!]
\centering
\includegraphics[width=0.8\columnwidth]{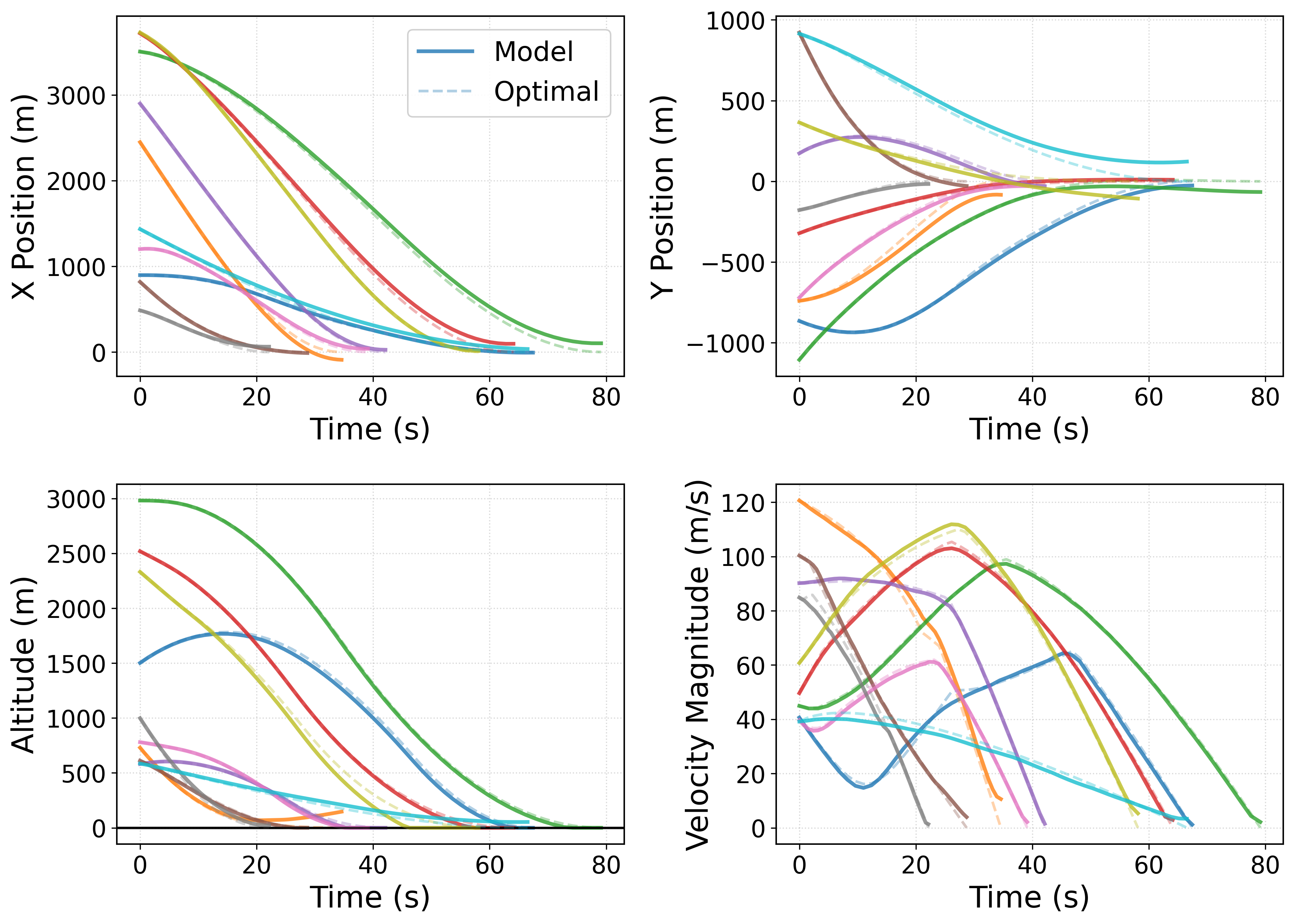}
\caption{Rocket landing state evolution over time: X position (top-left), Y position (top-right), altitude (bottom-left), and velocity magnitude (bottom-right). Solid lines show TRC predictions, dashed lines show optimal solutions.}
\label{fig:rocket_temporal}
\end{figure}

Figure~\ref{fig:rocket_temporal} shows the temporal evolution of key state components, illustrating the coordinated multi-dimensional control required for powered descent. The X and Y positions (top row) exhibit smooth convergence to zero, demonstrating effective lateral navigation that simultaneously guides the vehicle horizontally while managing vertical descent. The altitude profile (bottom-left) decreases in a near-linear fashion characteristic of fuel-optimal trajectories, closely tracking the optimal solutions throughout. The velocity magnitude (bottom-right) shows gradual deceleration with TRC predictions (solid) maintaining close agreement with optimal trajectories (dashed). The network successfully coordinates control across all seven state dimensions to achieve the terminal velocity constraint ($\|\mathbf{v}(t_f)\| \leq 1.0$~m/s) required for safe touchdown.

\begin{figure}[htb!]
\centering
\includegraphics[width=0.8\columnwidth]{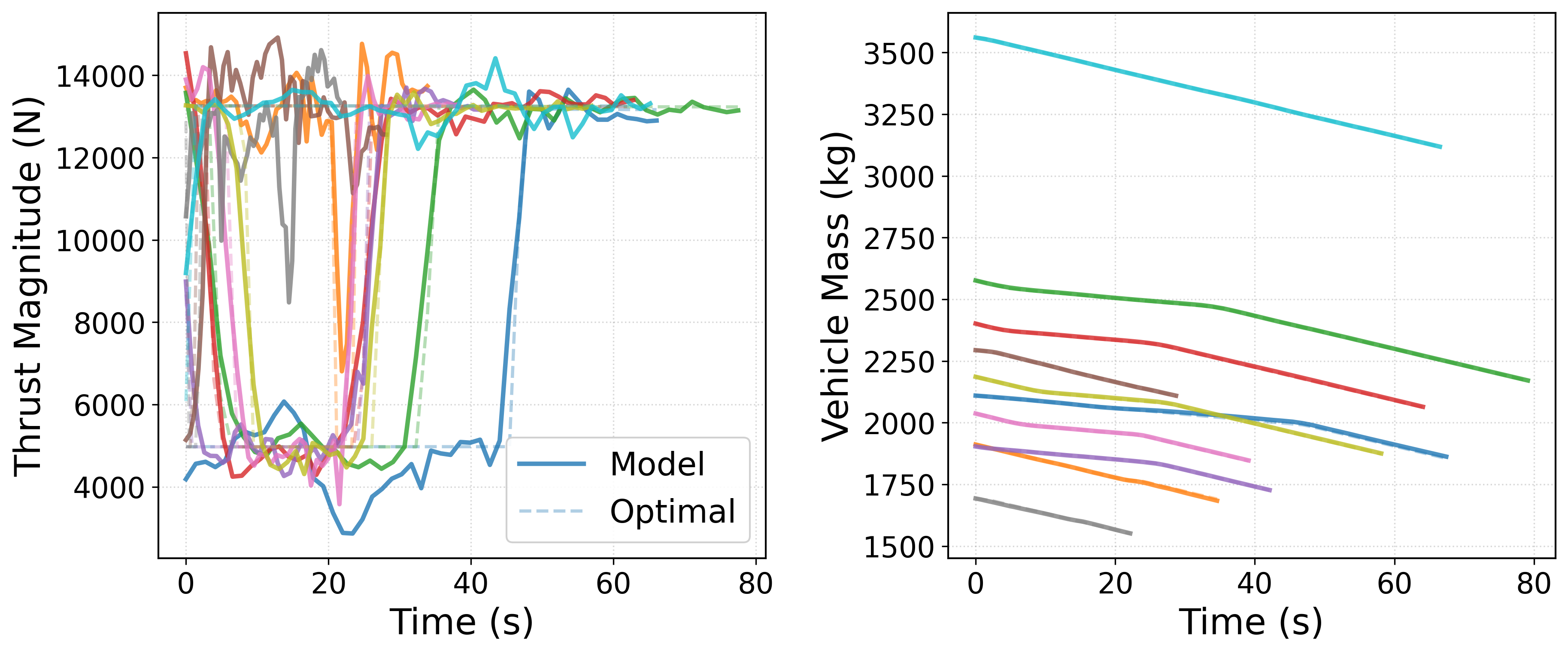}
\caption{Rocket landing control profile: thrust magnitude (left) and vehicle mass (right) over time. The thrust exhibits fuel-optimal bang-bang behavior, while mass decreases with propellant consumption.}
\label{fig:rocket_control}
\end{figure}

Figure~\ref{fig:rocket_control} presents the control profile and fuel consumption. The thrust magnitude (left) exhibits the bang-bang structure characteristic of fuel-optimal powered descent~\cite{acikmese2007convex}, alternating between maximum thrust ($T_{\max} = 13{,}000$~N) during initial deceleration and minimum thrust ($T_{\min} = 4{,}000$~N) during coast phases. Notably, TRC learns to reproduce this discontinuous switching behavior from optimal demonstrations despite using smooth neural network approximations, with process supervision encouraging each refinement iteration to reduce trajectory cost. The mass profile (right) shows propellant consumption over time, with TRC solutions closely tracking the optimal mass trajectories. The consistent mass evolution across different initial conditions indicates that TRC captures the fuel-efficient structure of the optimal policy, learning to balance deceleration requirements against propellant constraints.

\begin{figure}[htb!]
\centering
\includegraphics[width=\columnwidth]{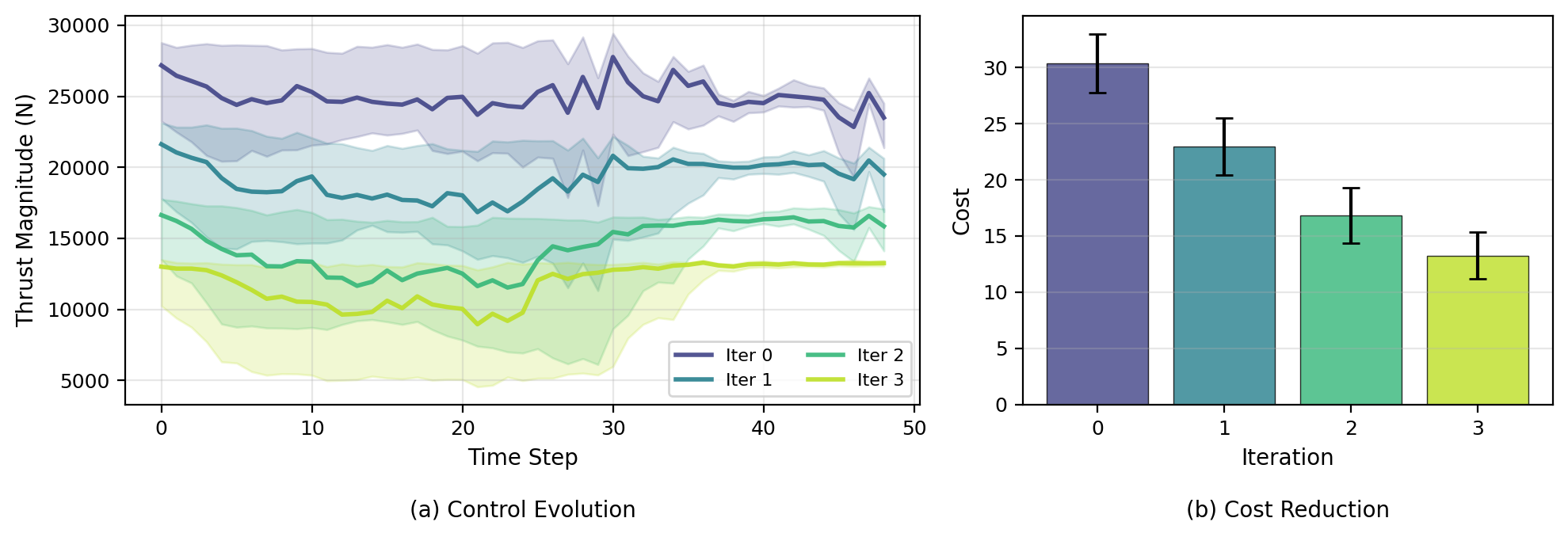}
\caption{Iterative refinement for powered descent guidance. (a) Thrust magnitude evolution across refinement iterations showing median (solid lines) with 25th-75th percentile bands. Controls converge from high-variance initial predictions (purple) to consistent fuel-optimal profiles (yellow). (b) Normalized fuel consumption (Eq.~\ref{eq:fuel_cost}) decreases monotonically across iterations, demonstrating that TRC learns to progressively reduce propellant usage.}
\label{fig:rocket_refinement}
\end{figure}

\begin{figure}[htb!]
\centering
\includegraphics[width=\columnwidth]{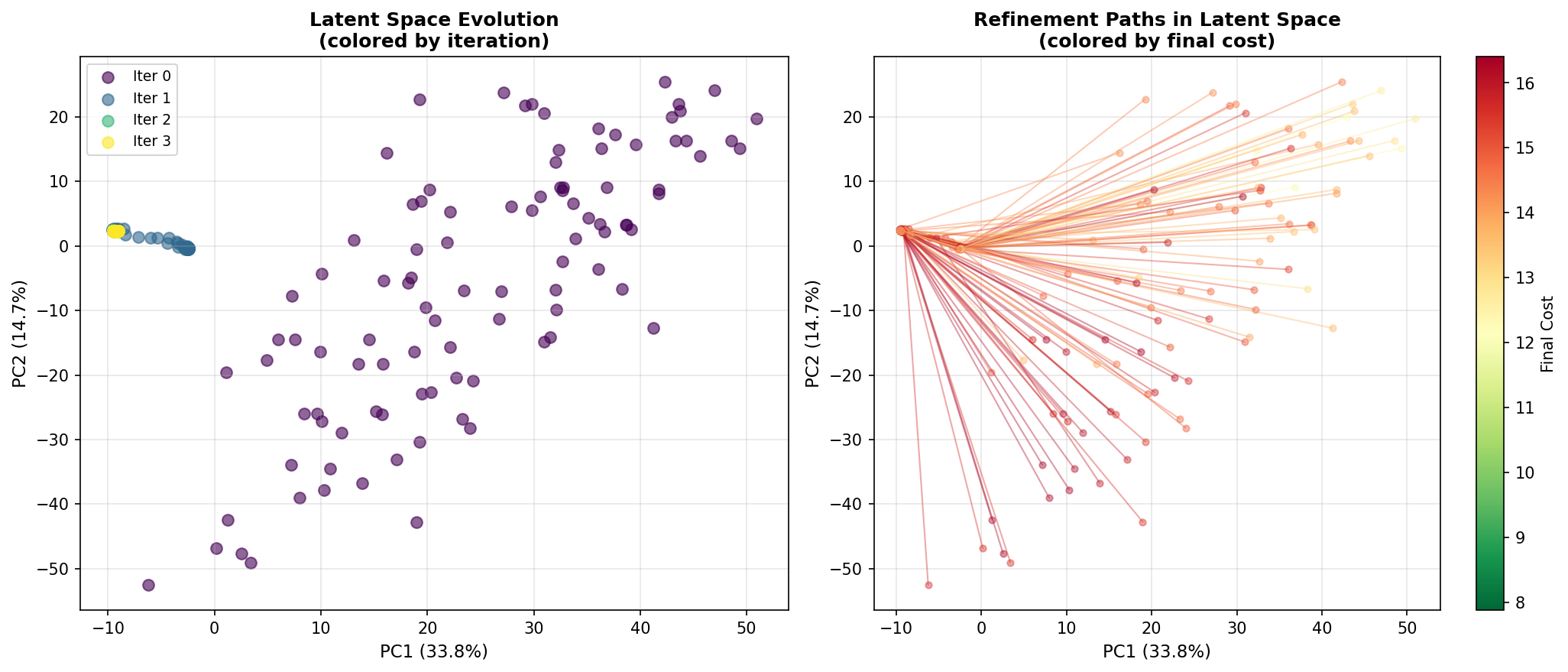}
\caption{Latent space evolution for powered descent guidance. (Left) Samples projected via PCA, colored by iteration. Scattered initial states (purple) converge to a compact region despite the high-dimensional control problem. (Right) Refinement trajectories colored by final cost show consistent flow toward low-cost solutions.}
\label{fig:rocket_latent}
\end{figure}

The iterative refinement analysis for powered descent is shown in Figures~\ref{fig:rocket_refinement} and~\ref{fig:rocket_latent}. Figure~\ref{fig:rocket_refinement} demonstrates that the refinement process is effective even for this high-dimensional control problem. The control evolution panel (a) shows the median thrust magnitude across test samples (solid lines) with 25th--75th percentile bands (shaded regions): initial estimates (purple, Iter~0) exhibit high thrust around 25,000~N with substantial variance, while successive iterations progressively reduce both the thrust magnitude and variance, converging toward fuel-efficient profiles near the operating bounds (yellow, Iter~3). The cost reduction panel (b) shows the mean fuel consumption (scaled by 1/10 for visualization) with standard deviation error bars across iterations. Cost decreases monotonically from approximately 30 (300~kg fuel) at iteration~0 to approximately 13 (130~kg fuel) at iteration~3, representing a 57\% reduction in propellant usage. This consistent improvement confirms that process supervision successfully teaches the refinement operator to produce increasingly fuel-efficient trajectories.

The latent space analysis in Figure~\ref{fig:rocket_latent} reveals a striking convergence phenomenon. The left panel shows latent states projected via PCA and colored by iteration: initial states (purple, Iter~0) are widely scattered across the principal component space, reflecting the diversity of descent scenarios with varying altitudes, velocities, and fuel states. After just one refinement iteration (teal, Iter~1), the points collapse dramatically into a tight cluster. By iteration~3 (yellow), all samples converge to essentially the same attractor point regardless of their initial conditions. The right panel visualizes individual refinement trajectories colored by final cost, showing all paths originating from scattered locations but flowing toward the same low-cost attractor region. This progressive collapse from diverse initial states to a universal representation demonstrates that TRC learns a fuel-optimal descent strategy that generalizes across initial conditions, rather than memorizing individual trajectories.

% ==============================================================================
% CONCLUSION (Section 6)
% ==============================================================================

\section{Conclusion}
\label{sec:conclusion}

This paper developed Tiny Recursive Control (TRC), a neural architecture that applies iterative refinement with weight sharing to continuous optimal control. The core insight is that capacity should come from iteration depth rather than parameter count. The empirical results demonstrate that compact networks (approximately 1.5M parameters), one to two orders of magnitude smaller than language model approaches, achieve near-optimal performance across two evaluation systems: Van der Pol oscillator and powered descent. Inference completes in under 10 ms on GPU with under 10 MB memory for weights, enabling deployment on embedded platforms where larger models are infeasible. The weight-sharing mechanism is central to this efficiency: the same reasoning blocks process every refinement iteration, so adding iterations increases computation but not memory. Beyond efficiency, the recursive architecture provides interpretability through inspectable intermediate control sequences, a crucial property for safety-critical aerospace applications.

Several limitations warrant acknowledgment and suggest future directions. TRC does not provide provable stability certificates; formal guarantees require future work on Lyapunov-based training losses or verification methods. The current approach requires access to optimal trajectories for training, and outputs fixed-length control sequences with constraint handling via clipping and penalty terms. Promising extensions include incorporating Lyapunov-based training for formal stability, developing differentiable barrier functions for explicit constraints, enabling adaptation to new dynamics through meta-learning, and validating real-world performance on embedded flight computers. Recursive neural reasoning represents a promising paradigm for efficient, interpretable control systems in aerospace and beyond.

% ==============================================================================
% ACKNOWLEDGMENTS
% ==============================================================================

\section*{Acknowledgments}
This work is supported by the Aerospace Corporation’s University Partnership Program.

% ==============================================================================
% REFERENCES
% ==============================================================================

\bibliography{references}

@misc{jolicoeurmartineau2024lessmore,
  title         = {Less is More: Recursive Reasoning with Tiny Networks},
  author        = {Alexia Jolicoeur-Martineau},
  year          = {2024},
  eprint        = {2510.04871},
  archiveprefix = {arXiv},
  primaryclass  = {cs.LG},
  url           = {https://arxiv.org/abs/2510.04871}
}

@book{anderson1971linear,
  author    = {Anderson, Brian D. and Moore, John B.},
  title     = {Linear Optimal Control},
  publisher = {Prentice-Hall},
  address   = {Englewood Cliffs, NJ},
  year      = {1971}
}

@book{lewis2012optimal,
  author    = {Lewis, Frank L. and Vrabie, Draguna and Syrmos, Vassilis L.},
  title     = {Optimal Control},
  edition   = {3},
  publisher = {Wiley},
  address   = {Hoboken, NJ},
  year      = {2012}
}

@book{camacho2013model,
  author    = {Camacho, Eduardo F. and Bordons, Carlos},
  title     = {Model Predictive Control},
  edition   = {2},
  publisher = {Springer-Verlag},
  address   = {London},
  year      = {2013}
}

@book{rawlings2017model,
  author    = {Rawlings, James B. and Mayne, David Q. and Diehl, Moritz M.},
  title     = {Model Predictive Control: Theory, Computation, and Design},
  edition   = {2},
  publisher = {Nob Hill Publishing},
  address   = {Madison, WI},
  year      = {2017}
}

@article{wu2024composing,
  author    = {Wu, Yinghan and Wang, Yizhuo and Zhang, Sirui and Wagle, Naren and Kobilarov, Marin and Sukhatme, Gaurav and Bezzo, Nicola},
  title     = {Composing {MPC} With {LQR} and Neural Network for Amortized Efficiency and Stable Control},
  journal   = {IEEE Transactions on Automation Science and Engineering},
  year      = {2024},
  publisher = {IEEE}
}

@article{celestini2024transformer,
  author    = {Celestini, Giulio and Gammelli, Daniele and Guffanti, Tommaso and D'Amico, Simone and Pavone, Marco},
  title     = {Transformer-based Model Predictive Control: Trajectory Optimization via Sequence Modeling},
  journal   = {IEEE Robotics and Automation Letters},
  volume    = {9},
  number    = {11},
  pages     = {9820--9827},
  year      = {2024},
  publisher = {IEEE}
}

@inproceedings{chen2018approximating,
  author    = {Chen, Steven and Saulnier, Kelsey and Atanasov, Nikolay and Lee, Daniel D. and Kumar, Vijay and Pappas, George J. and Morari, Manfred},
  title     = {Approximating explicit model predictive control using constrained neural networks},
  booktitle = {Proceedings of the American Control Conference (ACC)},
  pages     = {1520--1527},
  year      = {2018}
}

@article{hertneck2018learning,
  author  = {Hertneck, Michael and K{\"o}hler, Johannes and Trimpe, Sebastian and Allg{\"o}wer, Frank},
  title   = {Learning an approximate model predictive controller with guarantees},
  journal = {IEEE Control Systems Letters},
  volume  = {2},
  number  = {3},
  pages   = {543--548},
  year    = {2018}
}

@article{pomerleau1991efficient,
  author  = {Pomerleau, Dean A.},
  title   = {Efficient training of artificial neural networks for autonomous navigation},
  journal = {Neural Computation},
  volume  = {3},
  number  = {1},
  pages   = {88--97},
  year    = {1991}
}

@article{schulman2017proximal,
  author  = {Schulman, John and Wolski, Filip and Dhariwal, Prafulla and Radford, Alec and Klimov, Oleg},
  title   = {Proximal policy optimization algorithms},
  journal = {arXiv preprint arXiv:1707.06347},
  year    = {2017}
}

@inproceedings{haarnoja2018soft,
  author    = {Haarnoja, Tuomas and Zhou, Aurick and Abbeel, Pieter and Levine, Sergey},
  title     = {Soft actor-critic: Off-policy maximum entropy deep reinforcement learning with a stochastic actor},
  booktitle = {Proceedings of the International Conference on Machine Learning (ICML)},
  pages     = {1861--1870},
  year      = {2018}
}

@article{seeliger2022r2n2,
  author  = {Seeliger, Arne and others},
  title   = {A Recursively Recurrent Neural Network ({R2N2}) Architecture for Learning Iterative Algorithms},
  journal = {arXiv preprint arXiv:2211.12386},
  year    = {2022}
}

@inproceedings{carion2020end,
  author    = {Carion, Nicolas and Massa, Francisco and Synnaeve, Gabriel and Usunier, Nicolas and Kirillov, Alexander and Zagoruyko, Sergey},
  title     = {End-to-end object detection with transformers},
  booktitle = {Proceedings of the European Conference on Computer Vision (ECCV)},
  pages     = {213--229},
  year      = {2020}
}

@article{brohan2023rt2,
  author  = {Brohan, Anthony and others},
  title   = {{RT-2}: Vision-language-action models transfer web knowledge to robotic control},
  journal = {arXiv preprint arXiv:2307.15818},
  year    = {2023}
}

@article{jain2025multi,
  title   = {Multi-Phase Spacecraft Trajectory Optimization via Transformer-Based Reinforcement Learning},
  author  = {Jain, Amit and Rodriguez-Fernandez, Victor and Linares, Richard},
  journal = {arXiv preprint arXiv:2511.11402},
  year    = {2025}
}

@article{jain2025hamilton,
  title   = {A Hamilton-Jacobi Approach for Nonlinear Model Predictive Control in Applications with Navigational Uncertainty},
  author  = {Jain, Amit and Eapen, Roshan T and Singla, Puneet},
  journal = {arXiv preprint arXiv:2503.23603},
  year    = {2025}
}

@inproceedings{jain2023sparse,
  title        = {Sparse Approximate Hamilton-Jacobi Solutions for Optimal Feedback Control with Terminal Constraints},
  author       = {Jain, Amit and Eapen, Roshan and Singla, Puneet},
  booktitle    = {2023 62nd IEEE Conference on Decision and Control (CDC)},
  pages        = {1269--1274},
  year         = {2023},
  organization = {IEEE}
}

@book{bertsekas2012dynamic,
  title     = {Dynamic Programming and Optimal Control},
  author    = {Bertsekas, Dimitri P},
  volume    = {1},
  year      = {2012},
  publisher = {Athena Scientific}
}

@article{betts1998survey,
  title   = {Survey of numerical methods for trajectory optimization},
  author  = {Betts, John T},
  journal = {Journal of Guidance, Control, and Dynamics},
  volume  = {21},
  number  = {2},
  pages   = {193--207},
  year    = {1998}
}

@article{posadas2025action,
  title   = {Action Chunking with Transformers for Image-Based Spacecraft Guidance and Control},
  author  = {Posadas-Nava, Alejandro and Scorsoglio, Andrea and Ghilardi, Luca and Furfaro, Roberto and Linares, Richard},
  journal = {arXiv preprint arXiv:2509.04628},
  year    = {2025}
}

@inproceedings{lightman2023let,
  title     = {Let's verify step by step},
  author    = {Lightman, Hunter and Kosaraju, Vineet and Burda, Yuri and Edwards, Harrison and Baker, Bowen and Lee, Teddy and Leike, Jan and Schulman, John and Sutskever, Ilya and Cobbe, Karl},
  booktitle = {The Twelfth International Conference on Learning Representations},
  year      = {2023}
}

@inproceedings{briden2023tpdg,
  title     = {Improving Computational Efficiency for Powered Descent Guidance via Transformer-based Tight Constraint Prediction},
  author    = {Briden, Julia and Gurga, Trey and Johnson, Breanna and Cauligi, Abhishek and Linares, Richard},
  booktitle = {AIAA SciTech Forum},
  year      = {2024}
}

@article{acikmese2007convex,
  title   = {Convex programming approach to powered descent guidance for {Mars} landing},
  author  = {A{\c{c}}{\i}kme{\c{s}}e, Beh{\c{c}}et and Ploen, Scott R},
  journal = {Journal of Guidance, Control, and Dynamics},
  volume  = {30},
  number  = {5},
  pages   = {1353--1366},
  year    = {2007}
}

@article{zucchelli2025finetuned,
    title = {Fine-Tuned Language Models as Space Systems Controllers},
    author = {Zucchelli, Enrico and others},
    journal = {arXiv preprint arXiv:2501.16588},
    year = {2025},
    url = {https://arxiv.org/abs/2501.16588}
}

@article{carrasco2024finetuning,
    title = {Fine-tuning LLMs for Autonomous Spacecraft Control: A Case Study Using Kerbal Space Program},
    author = {Carrasco, Alejandro and Rodriguez-Fernandez, Victor and Linares, Richard},
    journal = {arXiv preprint arXiv:2408.08676},
    year = {2024},
    url = {https://arxiv.org/abs/2408.08676}
}

\end{document}